\documentclass[letterpaper]{article} 
\usepackage[submission]{aaai25}  
\usepackage{times}  
\usepackage{helvet}  
\usepackage{courier}  
\usepackage[hyphens]{url}  
\usepackage{graphicx} 
\usepackage{bm}
\usepackage{natbib}  
\usepackage{caption} 
\usepackage{algorithm}
\usepackage{algorithmic}
\usepackage{amsmath}
\usepackage{cuted}
\usepackage{newfloat}
\usepackage{listings}
\DeclareCaptionStyle{ruled}{labelfont=normalfont,labelsep=colon,strut=off} 
\floatstyle{ruled}
\newfloat{listing}{tb}{lst}{}
\floatname{listing}{Listing}
\title{TCR-GPT: Integrating Autoregressive Model and Reinforcement Learning for T-Cell Receptor Repertoires Generation}
\author {
    Yicheng Lin,
    Dandan Zhang,
    Yun Liu
}
\affiliations {
    School Of Basic Medical Sciences, Fudan University\\
    linyc20@fudan.edu.cn, zhangdandan3@hotmail.com, yliu39@fudan.edu.cn
}

\begin{document}

\maketitle

\begin{abstract}
T-cell receptors (TCRs) play a crucial role in the immune system by recognizing and binding to specific antigens presented by infected or cancerous cells. Understanding the sequence patterns of TCRs is essential for developing targeted immune therapies and designing effective vaccines. Language models, such as auto-regressive transformers, offer a powerful solution to this problem by learning the probability distributions of TCR repertoires, enabling the generation of new TCR sequences that inherit the underlying patterns of the repertoire.
We introduce TCR-GPT, a probabilistic model built on a decoder-only transformer architecture, designed to uncover and replicate sequence patterns in TCR repertoires. TCR-GPT demonstrates an accuracy of 0.953 in inferring sequence probability distributions measured by Pearson correlation coefficient. Furthermore, by leveraging Reinforcement Learning(RL), we adapted the distribution of TCR sequences to generate TCRs capable of recognizing specific peptides, offering significant potential for advancing targeted immune therapies and vaccine development. With the efficacy of RL, fine-tuned pretrained TCR-GPT models demonstrated the ability to produce TCR repertoires likely to bind specific peptides, illustrating RL’s efficiency in enhancing the model’s adaptability to the probability distributions of biologically relevant TCR sequences.

\end{abstract}

%

\section{Introduction}

T-cell receptors (TCRs) are integral components of the immune system, playing a crucial role in recognizing and responding to antigens. Each TCR is composed of unique sequences of amino acids, enabling T cells to identify specific pathogens and infected cells. The diversity and specificity of TCR sequences are vital for the adaptive immune response, allowing the immune system to target a vast array of antigens with precision
 \cite{ref_17_Singh,ref_18_van}.
Understanding the repertoire of TCR sequences is essential for advancing immunological research and developing targeted therapies.

Text-generation models, inspired by natural language processing techniques
 \cite{ref_04_Chung,ref_08_Hochreiter,ref_19_Vaswani}
, can potentially offer a transformative approach to this task. By building a probability distribution over TCR sequences, and even more, tailoring this distribution to generate TCR sequences of specific desired functionality, text-generation models can not only enhance our understanding of TCR diversity but also has practical applications in designing vaccines and immunotherapies.

In the realm of auto-regressive text generation, decoder-only transformers have become a cornerstone technology
. This architecture processes input sequences to generate text in an auto-regressive manner, predicting the next token in a sequence based on the tokens it has previously generated. At each step, the model takes the current sequence of tokens as input, processes it through multiple layers of self-attention and feed-forward networks, and produces a probability distribution over the next possible tokens. The model then samples from this distribution to select the next token, appends it to the input sequence, and repeats the process until the desired output length is achieved or a stopping condition is met. Decoder-only transformers like GPT (Generative Pre-trained Transformer) 
\cite{ref_03_Brown,ref_12_Radford2018,ref_13_Radford2019} 
 have demonstrated remarkable performance in various natural language processing tasks, from completing sentences and paragraphs to generating entire articles or dialogues, thus setting new benchmarks for text generation capabilities.

Reinforcement Learning (RL) involves training agents to maximize long-term rewards by interacting with complex environments. This method has been applied across diverse fields such as game playing \cite{ref_23_RL_gaming}, robotics \cite{ref_22_RL_robotics}, drug discovery \cite{ref_24_RL_drug,ref_25_RL_drug,ref_26_RL_drug,ref_27_RL_drug,ref_28_RL_drug}, and more. Reinforcement learning has also emerged as an effective alternative training approach to enhance the performance of generative models, offering adaptable goals via its reward function, unlike the fixed distribution modeling objectives found in both supervised and unsupervised learning \cite{ref_29_RL_generative}. 

In this study, we introduce TCR-GPT, a probabilistic model based on a decoder-only transformer architecture designed to capture underlying sequence patterns in TCR repertoires and hence generate TCR sequences based on the learned probability distribution. And through reinforcement learning, we tailored the distribution of TCR sequences to generate TCRs that can recognize specific peptides, which holds great promise for advancing targeted immune therapy and vaccine design. TCR-GPT demonstrates greater accuracy in inferring sequence probability distributions compared to existing models, with average performance 0.953 as measured by the correlation coefficient. We further verified that TCR-GPT can efficiently adapt to different TCR sub-repertoires and provide learnable features for TCR sequences. After that, we demonstrate the effectiveness of RL in fine-tuning pretrained TCR-GPT models to generate TCR repertoires likely to bind specific peptides. Using PanPep \cite{ref_07_Gao}, a robust peptide-TCR binder prediction tool, we iteratively refine TCR-GPT through RL to generate peptide-specific TCR repertoires, showcasing RL’s efficiency in significantly enhancing the model’s ability to adapt to the underlying probability distribution of biologically relevant TCR sequences.

\section{Related work}

Traditional methods for modeling TCR sequence patterns calculate the probability of a TCR sequence as the product of the selection factor and the generation probability, which are inferred separately from the processes of TCR selection and generation. A notable example of this approach is soNNia \cite{ref_09_Isacchini}.

Trials have also been conducted using variational autoencoders (VAEs) parameterized by deep neural networks to model TCR repertoires \cite{ref_05_Davidsen}
. However, it requires TCR sequences to be padded to a fixed length, which can introduce noise into the original data and obscure valuable information about the diversity of sequence lengths.

Inspired by natural language processing and auto-regressive text generation models, TCRpeg \cite{ref_10_Jiang} 
employs an auto-regressive framework to infer the probability of TCR sequences in an end-to-end manner. TCRpeg utilizes a deep autoregressive architecture with gated recurrent unit (GRU) layers \cite{ref_04_Chung}
 to effectively model the probability of TCR sequences, offering a streamlined alternative to the above trials.

Transformer models, introduced by Vaswani et al. in 2017, have revolutionized the field of natural language processing (NLP) \cite{ref_19_Vaswani}
. Unlike previous sequence-to-sequence models that relied heavily on recurrent neural networks (RNNs) \cite{ref_04_Chung,ref_08_Hochreiter,ref_14_Schulman}
 and convolutional neural networks (CNNs) \cite{ref_21_CNN}, transformers leverage self-attention mechanisms to process input data in parallel, allowing for greater efficiency and scalability. The core innovation of the transformer architecture is its ability to capture long-range dependencies within the data, which is crucial for understanding and generating coherent text. A notable example that combines transformer and auto-regressive models is the GPT (Generative Pre-trained Transformer) series \cite{ref_03_Brown,ref_12_Radford2018,ref_13_Radford2019}. These models generate text by predicting each word or token based on the preceding sequence, utilizing the attention mechanism inherent in transformer decoders to create coherent and contextually relevant text.

 Reinforcement learning has demonstrated its efficacy as a robust method for training generative models, especially in NLP tasks \cite{ref_29_RL_generative}. Reinforcement Learning with Human Feedback (RLHF) \cite{ref_30_RLHF,ref_31_RLHF,ref_32_RLHF,ref_33_RLHF,ref_34_RLHF} is a representative technique where reinforcement learning is used to fine-tune language models based on human-provided feedback. In the context of large language models (LLMs) like GPT, RLHF helps improve the model’s performance by aligning its outputs with human preferences, leading to more contextually appropriate responses.

\section{Methods}
\subsection{Autoregressive generative transformer for TCR sequences}
Given the critical role of the CDR3-$\beta$ chain in TCR antigen recognition \cite{ref_17_Singh,ref_18_van}
, we utilized the CDR3-$\beta$ sequence as a representation of the entire TCR. 
We developed an autoregressive model, TCR-GPT, to estimate the probability of a TCR sequence $\mathbf{x}$ as $p(\mathbf{x}|\bm{\theta})$, where $\bm{\theta}$ represents the parameters of a decoder-only transformer model. 
This probability $p(\mathbf{x}|\bm{\theta})$ is computed by the autoregressive likelihood, which involves calculating the product of conditional probabilities over sequential residues of length $L$:

\begin{equation}\label{EQ1}
    p\left(\mathbf{x}\middle|\bm{\theta}\right)=p\left(x_1\middle|\bm{\theta}\right)\prod_{i=2}^{L}{p\left(x_i\middle| x_1,x_2,\ldots,x_{i-1};\bm{\theta}\right)}
\end{equation}

Our model employs a decoder-only transformer equipped with attention mechanisms to capture interactive relationships among residues, thereby formulating the autoregressive likelihood efficiently.
After embedding layers transform the discrete representation of each amino acid in batches of TCR sequences into continuous vectors of dimensionality 32, a total of 8 attention heads, each with 6 self-attention layers, collectively capture the features.
Specifically, let $\mathbf{X}\in[0,1,2,...,22]^{B\times L}$ denote a batch of $B$ TCRs of maximum length $L$.
Each element in $\mathbf{X}$ is an integer representing an amino acid (a total of 20 different amino acids), as well as special tokens such as $<\mathbf{SOS}>$ (Start of Sequence), $<\mathbf{EOS}>$ (End of Sequence), and padding token.
After embedding, $\mathbf{X}$ is transformed to a continuous tensor $E\in R^{B\times L\times d}$, where each vector of dimension $d=32$ represents the embedding of a single token.
The multi-head attention layers, along with the linear layers after multi-head feature concatenation then transform $E$ to features $Z\in R^{B\times L\times d}$. 
These features are then processed by a position-wise feedforward network to output the probability distribution over all 20 possible amino acids at each position. The overall architecture of TCR-GPT is depicted in Figure \ref{fig1}A. We set the negative log likelihood as the loss function as the following:

\begin{equation}\label{EQ2}
    L=-\sum_{n=1}^{N}\sum_{i=1}^{L}{logp_i}
\end{equation}

$N$ is the batch size and $p_i$ is the output probability of the model on the right amino acid in $i$-th amino acid position.

\newpage

\begin{strip}
\centering
\includegraphics[width=0.85\textwidth]{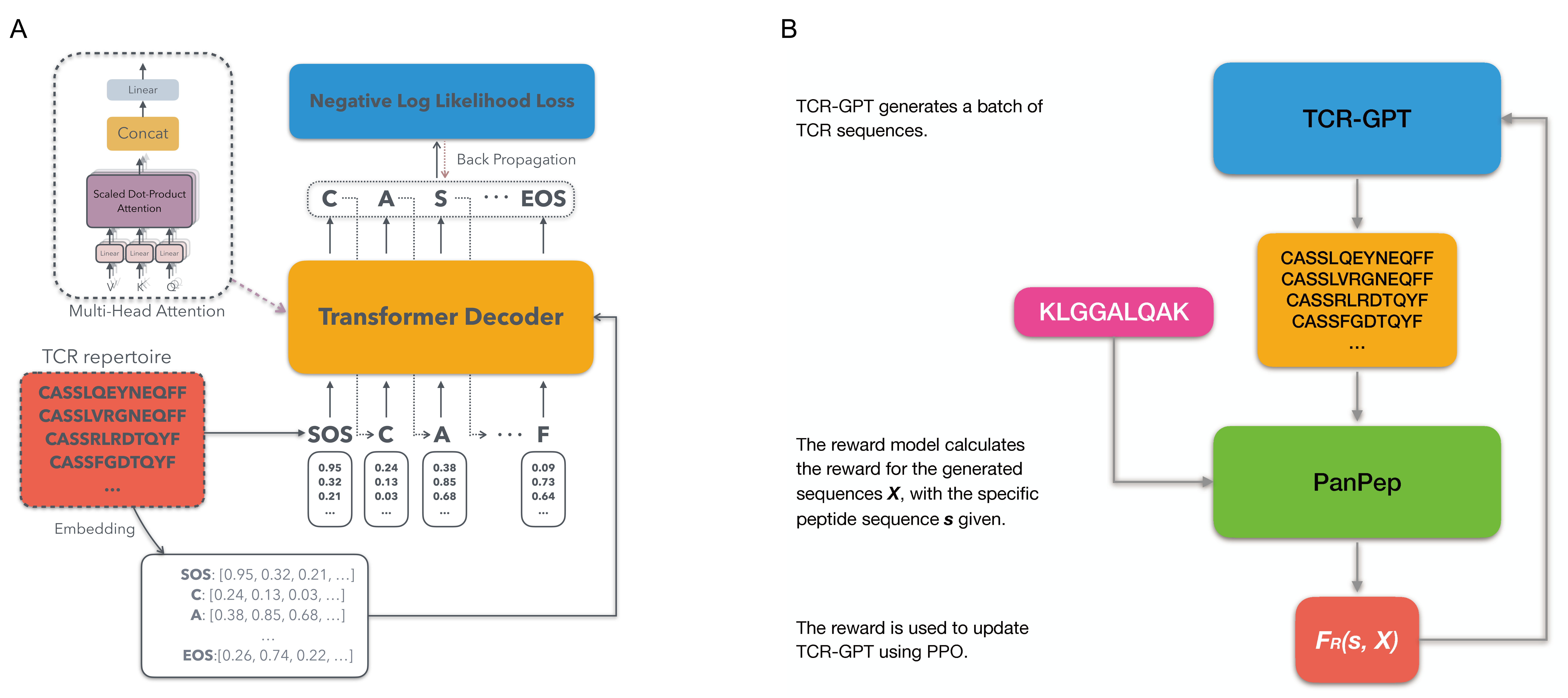} 
\captionof{figure}{(A) The overall architecture of TCR-GPT. (B) Main workflow of peptide-specific RL for TCR-GPT.}
\label{fig1}
\end{strip}

We trained the model on a dataset containing 50 million TCRs collected from a large cohort of 743 individuals, as described by Emerson et al 
\cite{ref_06_Emerson}
.



\subsection{Jensen-Shannon Divergence}
We use Jensen-Shannon divergence ($D_{js}(r^i,r^j)$) to measure the ability of the trained model to distinguish two sub-repertoires $r^i$ and $r^j$, which is formulated as:

\begin{equation}\label{EQ3}
    D_{js}(r^i,r^j)=\frac{1}{2}KL(P_{infer}^i||M)+\ \frac{1}{2}KL(P_{infer}^j||M)
\end{equation}

where $P_{infer}^i$ and $P_{infer}^j$ are probabilities computed from TCR-GPT trained on sub-repertoires $r^i$ and $r^j$, 
$M=\frac{1}{2}(P_{infer}^i+P_{infer}^j)$, 
and 
$KL(P||Q)$ denotes the Kullback-Leibler divergence. A higher $D_{js}(r^i,r^j)$ indicates greater discrepancy between sub-repertoires.

For the scenario where we utilize Jensen-Shannon divergence to assess the difference of the inferred likelihood from the model $P_{infer}(\mathbf{x})$ and the observed frequencies $P_{data}(\mathbf{x})$, we simply calculate $D_{js}(P_{infer},P_{data})$ as the following:


\begin{equation}
	\left.
    \begin{array}{r}
 D_{js}(P_{infer},P_{data})=\frac{1}{2}KL(P_{infer}(\mathbf{x})||M) \\
+\frac{1}{2}KL(P_{data}(\mathbf{x})||M) \\
    \end{array}
    \right.
\end{equation}

\noindent where $M=\frac{1}{2}(P_{infer}(\mathbf{x})+P_{data}(\mathbf{x}))$.

\subsection{Using TCR-GPT to generate TCR sequences}
With the TCR-GPT model trained, we employed a straightforward sampling method to generate new TCR sequences. Initially, we input the start token ($<\mathbf{SOS}>$) into the model and then sampled the amino acid for the next position from the resulting probability distribution. This sampled amino acid was appended to the previous tokens, and the process was repeated iteratively to generate each subsequent amino acid. This approach can be summarized by the following formula:

\begin{equation}\label{EQ4}
    A_t=P(A|A_{0:t-1};\bm{\theta})
\end{equation}

Here, $A_t$ represents the amino acid token at position $t$, and $\bm{\theta}$ denotes the parameters of the TCR-GPT model. The generation process concludes when the $<\mathbf{EOS}>$ (End Of Sequence) token is generated. At this point, all the generated amino acids are collected to form the complete TCR sequence.

\subsection{Using the features from TCR-GPT for classification tasks}
Upon training the TCR-GPT model, the features produced by the multi-head attention module can encode valuable information from the sequence input. We developed an additional classification network that utilizes these features for a downstream classification task. 
Specifically, let $\mathbf{X}\in[0,1,2,...,22]^{B\times L}$ represent a batch of $B$ TCRs with a maximum length of $L$.
The features extracted from the multi-head attention module are denoted as $Z\in R^{B\times L\times d}$, where the feature of the amino acid at each position in the $L$-length sequence is a $d$-dimensional tensor.
We then flattened the feature $Z$ to $Z_{flatten}\in R^{B\times q}$, where each TCR sequence is represented as a $q$-dimensional tensor. We constructed a fully connected neural network for each classification task, which consists of three linear layers that can be formulated as follows:

\begin{equation}\label{EQ5}
    Z_0=Z_{flatten}
\end{equation}

\begin{equation}\label{EQ6}
    Z_l=\sigma(W_lZ_{l-1}+b_{l-1}), l\in\left\{1,2\right\}
\end{equation}

\begin{equation}\label{EQ7}
    P=softmax(W_3Z_2+b_3)
\end{equation}

\noindent here, $W_l$ and $b_l$ represent the parameters of the $l$-th layer of the fully connected neural network, and $\sigma$ denotes the activation function. The final layer, combined with the softmax operation, produces the predicted probability distribution of the classes.

\subsection{Fine-tuning TCR-GPT with RL to generate peptide-specific TCR repertoires}
To generate TCR repertoires capable of binding to specific peptides, we employ peptide-specific RL to integrate knowledge from high-performance peptide-TCR binding predictors into our pretrained TCR-GPT model. PanPep
 \cite{ref_07_Gao}
, a versatile and robust model for evaluating peptide-TCR binding affinity, leverages meta-learning and neural Turing machine concepts to achieve superior performance in both few-shot and zero-shot scenarios. We adopt PanPep as the reward model in our reinforcement learning framework. During the reinforcement learning iteration, PanPep rates the generated TCR sequences from TCR-GPT based on their binding affinity to the given peptide.

Specifically, with given peptide denoted as $\mathbf{s}$, let the reward for the generated TCR sequences from the TCR-GPT model be denoted as $F_R(\mathbf{s},\mathbf{X})$. Here, PanPep serves as the reward function $F_R$ for the generated TCR sequences. We constructed another transformer network parameterized by $\bm{\varphi}_\mathbf{k}$ at the $k$-th iteration as the critic $V^{\bm{\varphi}_\mathbf{k}}$, which shares the same architecture as the TCR-GPT model (the actor in this scenario). However, the critic’s output is a single value representing the predicted return for each TCR sequence entry. Using PPO \cite{ref_14_Schulman}
 as the reinforcement learning algorithm, we formulate the objective function as 



\begin{equation}\label{EQ8}
\left.
\begin{array}{l}
    L\left(\mathbf{s},\mathbf{X},\bm{\theta}_\mathbf{k},\bm{\theta}\right) \\
    =\min(\frac{p_{\bm{\theta}}\left(\mathbf{X}\middle|\mathbf{s}\right)}{p_{\bm{\theta}_\mathbf{k}}\left(\mathbf{X}\middle|\mathbf{s}\right)}A^{p_{\bm{\theta}_\mathbf{k}}}\left(\mathbf{s},\ \mathbf{X}\right),g(\epsilon{,A}^{p_{\bm{\theta}_\mathbf{k}}}\left(\mathbf{s},\ \mathbf{X}\right)))

\end{array}
\right.
\end{equation}


where

\begin{equation}\label{EQ8-2}
    g(\epsilon,A) = \left\{
    \begin{array}{c}
      (1+\epsilon )A,\quad A\ge 0  \\  
      (1-\epsilon )A,\quad A<  0   
    \end{array}
    \right.
\end{equation}

\begin{equation}\label{EQ9}
    A^{p_{\bm{\theta}_\mathbf{k}}}(\mathbf{s},\ \mathbf{X})=F_R\left(\mathbf{s},\mathbf{X}\right)-V^{\varphi_k}(\mathbf{X})
\end{equation}

The advantage $A^{p_{\bm{\theta}}}(\mathbf{s},\ \mathbf{X})$ is the difference between reward $F_R\left(\mathbf{s},\mathbf{X}\right)$ and the value $V^{\varphi_k}(\mathbf{X})$, the predicted reward of TCR sequence $\mathbf{X}$ by the critic network parameterized by $\bm{\varphi}$ in the $k$-th iteration. $\epsilon$ is a hyperparameter that we set at 0.2 as used in the previous paper
 \cite{ref_14_Schulman}
. At $k$-th iteration, we update the parameters $\bm{\theta}$ of TCR-GPT, and the parameters $\bm{\varphi}$ of the crtic in the following formula:

\begin{equation}\label{EQ10}
    \bm{\theta}_{k+1}=\mathop{\arg\max}\limits_{\bm{\theta}}{\frac{1}{|D_k|}\sum_{\mathbf{X}\in D_k} L\left(\mathbf{s},\mathbf{X},\bm{\theta}_\mathbf{k},\bm{\theta}\right)}
\end{equation}

\begin{equation}\label{EQ11}
    \bm{\varphi}_{k+1}=\mathop{\arg\min}\limits_{\bm{\varphi}}{\frac{1}{\left|D_k\right|}}\sum_{\mathbf{X}\in D_k}[V^{\varphi k}(\mathbf{X})-F_R(\mathbf{s},\mathbf{X})]^2
\end{equation}

Where $D_k$ is a batch of TCR sequences collected using TCR-GPT model during the $k$-th iteration. The peptide-specific RL workflow can be visualized in Figure \ref{fig1}B.


\section{Experiments}
\subsection{TCR-GPT infers the probability distribution of TCR repertoires with high accuracy}
To evaluate the performance of TCR-GPT, we first applied it to infer the probability distribution of TCR-$\beta$ CDR3 sequences and compared its performance with two other algorithms, soNNia
 \cite{ref_09_Isacchini}
 and TCRpeg
 \cite{ref_10_Jiang}
 . We constructed a universal TCR repertoire by pooling the CDR3 sequences from a large cohort of 743 individuals
 \cite{ref_06_Emerson}
 . This comprehensive dataset provided a robust foundation for training and testing our model as well as the comparative algorithms. To assess the accuracy of these three methods, we randomly divided the universal repertoire into training and testing sets using a 50:50 ratio, the same strategy employed by TCRpeg. This approach ensured a fair and consistent comparison across the different models.
 
\begin{figure}[h]
\centering
\includegraphics[width=0.99\columnwidth]{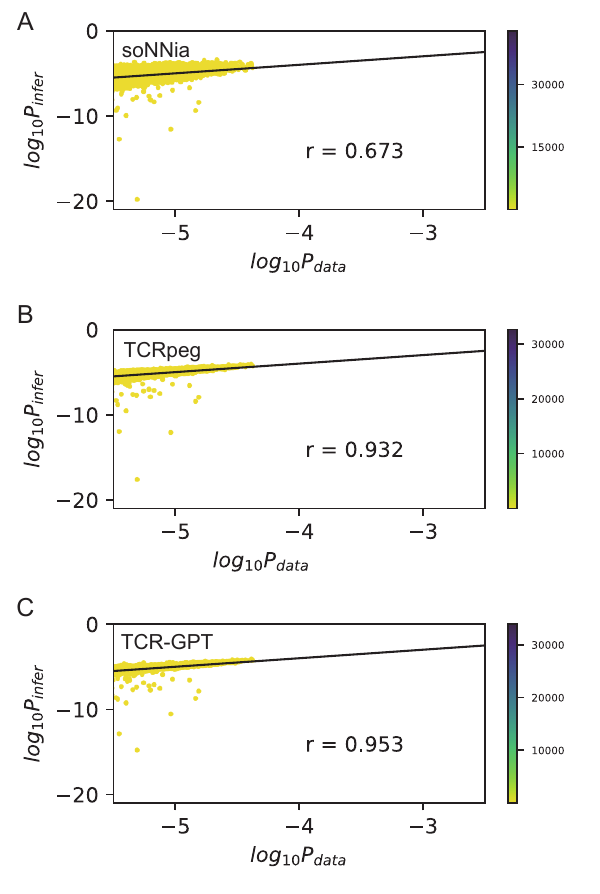} 
\caption{Performance comparison of TCR-GPT, soNNia and TCRpeg algorithms. A-C. The scatter plot of actual ($P_{data}$) versus inferred probability ($P_{infer}$) for soNNia (A), TCRpeg (B) and TCR-GPT (C) using test dataset from universal TCR repertoire. The corresponding Pearson correlation coefficients are displayed for each plot.}
\label{fig3}
\end{figure}

We measured the concordance between the inferred ($P_{infer}(\mathbf{x})$) and the actual probability distributions ($P_{data}(\mathbf{x})$) by Pearson correlation coefficients. TCR-GPT demonstrated greater concordance, achieving a Pearson correlation coefficient of $r=0.953$, compared to soNNia ($r=0.673$) and TCRpeg ($r=0.932$) (Figure \ref{fig3}A-C). These results indicate that TCR-GPT has the highest accuracy among the three methods, as reflected by its closer alignment with the actual probability distributions.

In addition to Pearson correlation coefficients, we utilized the Jessen-Shannon divergency ($D_{js}$, Methods) to assess the difference between $P_{infer}(\mathbf{x})$ and $P_{data}(\mathbf{x})$. The $D_{js}$ statistics provides a measure of divergence between two probability distributions, with a higher $D_{js}$ indicating a greater difference, and consequently lower model accuracy. The $D_{js}$ statistic further confirmed the higher accuracy of TCR-GPT, which had a $D_{js}$ value of 0.031. In contrast, soNNia and TCRpeg had $D_{js}$ values of 0.131 and 0.039, respectively. These results underscore the effectiveness of TCR-GPT in accurately inferring the probability distribution of TCR-$\beta$ CDR3 sequences.

According to the comparisons above, we conclude that TCR-GPT outperformed soNNia and TCRpeg in both measures of accuracy, making it the most reliable method among the three for inferring the probability distribution of TCR-$\beta$ CDR3 sequences. The superior performance of TCR-GPT, as evidenced by both Pearson correlation coefficients and Jensen-Shannon divergence, highlights its potential as a powerful tool for TCR repertoire analysis.

\subsection{TCR-GPT captures specific features of TCR repertoires efficiently}
We further investigated TCR-GPT’s ability to learn the specific probability distributions of different TCR repertoires to compare their properties from a probabilistic perspective. CD4 and CD8 T cells have distinct roles in adaptive immunity across various tissues, which may be reflected by their unique TCR repertoires. Utilizing a dataset
 \cite{ref_10_Jiang,ref_15_Seay}
 comprising three different cell types (CD8 cytotoxic T lymphocytes(CD8+), CD4 T help cells(Tconv) and CD4 Treg cells(Treg)) collected from three tissues (spleen, pancreatic draining lymph modes [pLN] or inguinal ‘irrelevant’ lymph nodes [iLN]), we compared the discrepancy among these nine sub-repertoires.

\begin{figure}[h]
\centering
\includegraphics[width=0.99\columnwidth]{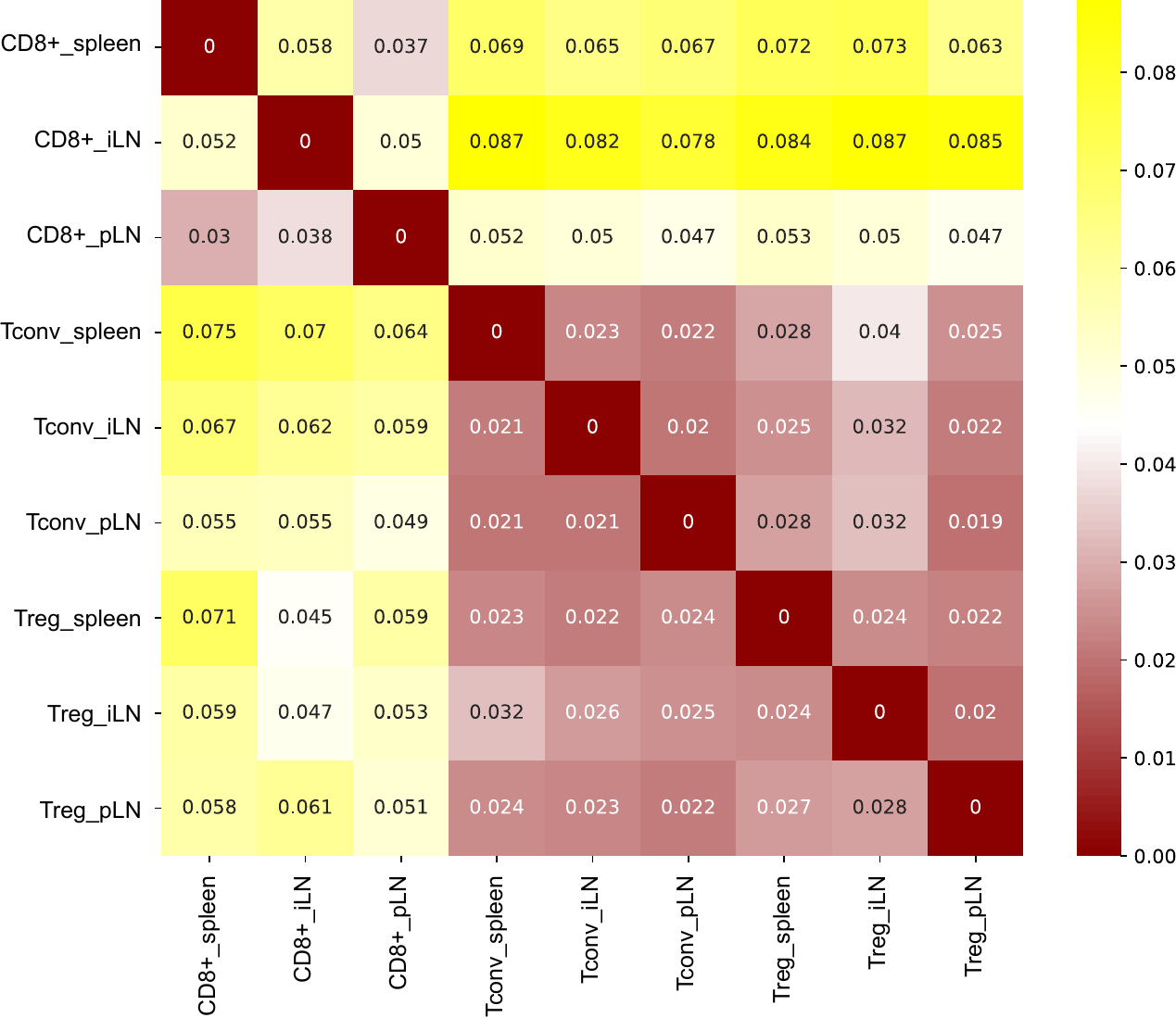} 
\caption{The heatmap of Jensen-Shannon divergence ($D_{js}$) between pairwise sub-repertoire probability distribution inferred by TCR-GPT.}
\label{fig4}
\end{figure}

We used TCR-GPT to construct a probabilistic generative model for each of these nine TCR sub-repertoires. For each sub-repertoire, we inferred its probability using nine models: one trained from its own TCR repertoire (denoted as $P$) and eight from the remaining TCR repertoires (denoted as $Qs$). To estimate the distance between each pair of sub-repertoires, we calculated the Jessen-Shannon divergency ($D_{js}$) for $P$ and each $Q$. The $D_{js}$ matrix revealed an interesting pattern: sub-repertoires of the same cell type were more similar to each other than to those from different cell types, and CD4 T cell subtypes exhibited greater similarity to each other than to CD8 T cells (Figure \ref{fig4}).

Given the fundamental role of the CDR3 sequence in determining T-cell function through antigen recognition (PMID: 28923982), these results suggest that the same cell subtypes with similar CDR3 repertoire performed analogous functions across different tissues.

Moreover, the higher similarity observed among CD4 T cell subtypes compared to CD8 T cells aligns with the distinct functional roles that CD4 and CD8 T cells play in the immune system. CD4 T cells are primarily involved in helper or regulator functions, facilitating the activation or inhibition and coordination of other immune cells, while CD8 T cells are chiefly responsible for cytotoxic activities, directly targeting and eliminating infected or malignant cells.

This probabilistic comparison of TCR repertoires across different cell types and tissues enhances our understanding of the adaptive immune system’s complexity. It provides insights into how specific TCR repertoires are tailored to meet the unique functional demands of different T cell subtypes in various tissue environments. Such knowledge is invaluable for advancing immunotherapy and vaccine development, where targeted manipulation of TCR repertoires could lead to more effective and precise treatments.

\subsection{Classification of cancer-associated TCRs and SARS-CoV-2 epitope-specific TCRs using features from TCR-GPT}
During the process of TCR probability inference, TCR-GPT yields features for each TCR sequence. Unlike predefined embeddings, the features provided by TCR-GPT are generated in a learnable manner, allowing the model to adaptively capture the underlying patterns within the TCR sequences.
\newpage

\begin{strip}
\centering
\includegraphics[width=0.9\textwidth]{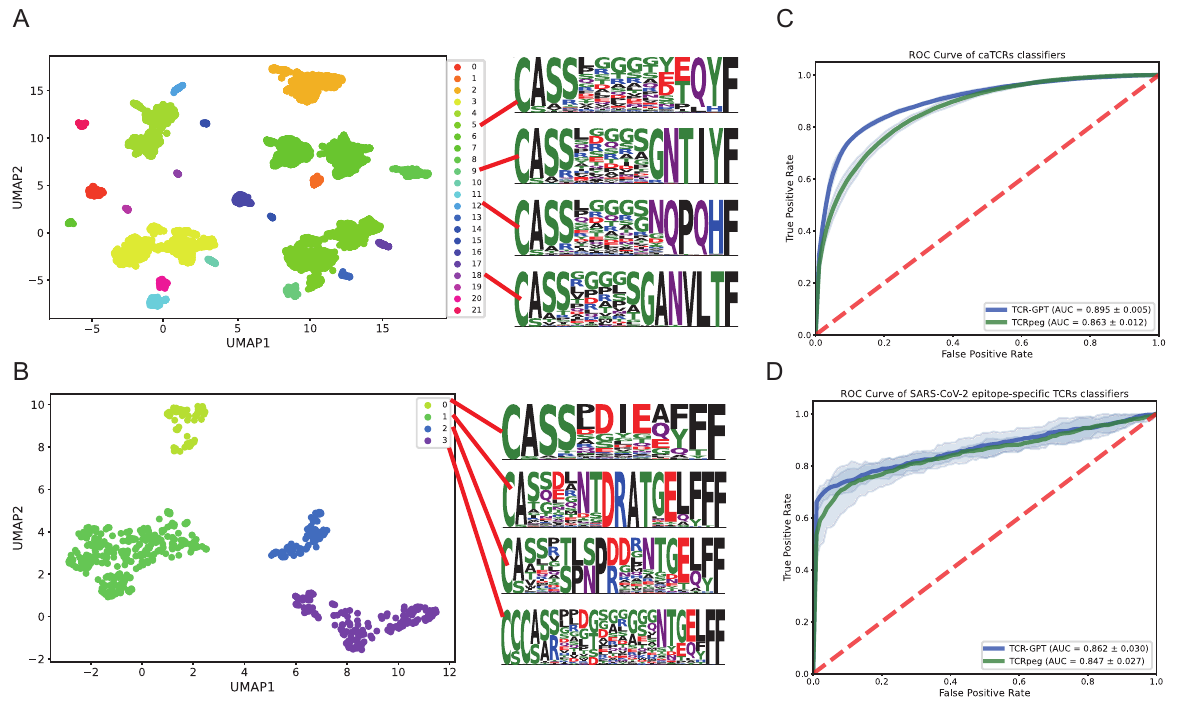} 
\captionof{figure}{UMAP Visualization and Classification Performance of TCR-GPT. (A, B) UMAP plots of features learned by TCR-GPT trained on caTCRs (A) and SARS-TCRs (B), along with motif logos of selected clustered TCR sequences. (C, D) Area under curve (AUC) for classifiers predicting caTCRs (C) and SARS-TCRs (D) trained with TCR-GPT.}
\label{fig5}
\end{strip}

\noindent To depict the features of TCR sequences, we evaluated cancer-associated TCRs (caTCRs) from Beshnova et al \cite{ref_02_Beshnova}
($N\sim43,000$) and SARS-CoV-2 epitope (YLQPRTFLL) specific TCRs (SARS-TCRs) from VDJdb database
 \cite{ref_16_Shugay}
 ($N=683$). We trained TCR-GPT using these two TCR repertoires separately to obtain the representative feature vectors of each sequence.

\begin{figure}[h]
\centering
\includegraphics[width=0.85\columnwidth]{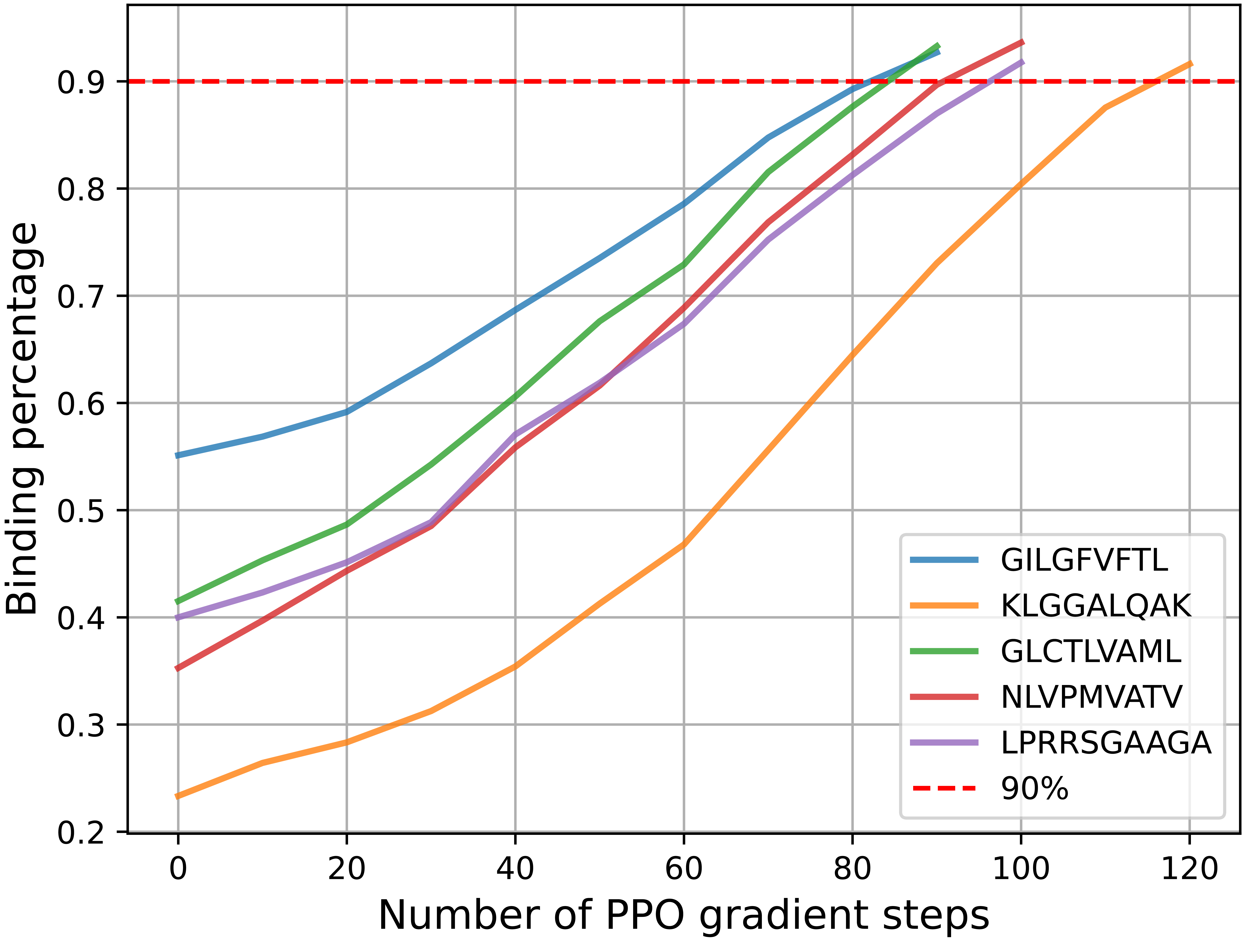} 
\caption{Binding percentage of generated TCRs with specific peptide sequences increases with the number of PPO gradient steps.}
\label{fig6}
\end{figure}

The Uniform Manifold Approximation and Projection (UMAP) dimensionality reduction applied to the feature space revealed explicit clusters in the 2D map (Figure \ref{fig5}A, B). Sequences within the same cluster shared a similar motifs, suggesting that the features generated by TCR-GPT could accurately represent the TCR sequences.

To further illustrate the effectiveness behind TCR-GPT’s features of TCRs, we performed classification(Methods) of caTCRs and SARS-TCRs from control (negative) TCRs sampled from the universal TCR repertoire mentioned above. We extensively compared TCR-GPT with TCRpeg in both caTCR and SARS-TCR classification tasks. Using five-fold cross-validation, TCR-GPT exhibited more stable and accurate performance than TCRpeg in caTCRs classification task, achieving Aera Under the Curve (AUC) values of 0.895±0.005 for TCR-GPT and 0.863±0.012 for TCRpeg (Figure \ref{fig5}C). 

The results from the SARS-TCR classification task also support the superior performance of TCR-GPT, with AUCs of 0.862±0.030 compared to 0.847±0.027 for TCRpeg (Figure \ref{fig5}D). The consistent performance of TCR-GPT across different classification tasks indicates its robustness and generalizability in modeling TCR sequences. The higher AUC values for TCR-GPT suggest that it can more effectively capture the relevant features that distinguish specific TCRs from the general repertoire, thereby enhancing its predictive accuracy.
\newpage

\begin{strip}
    \centering
    \includegraphics[width=0.8\textwidth]{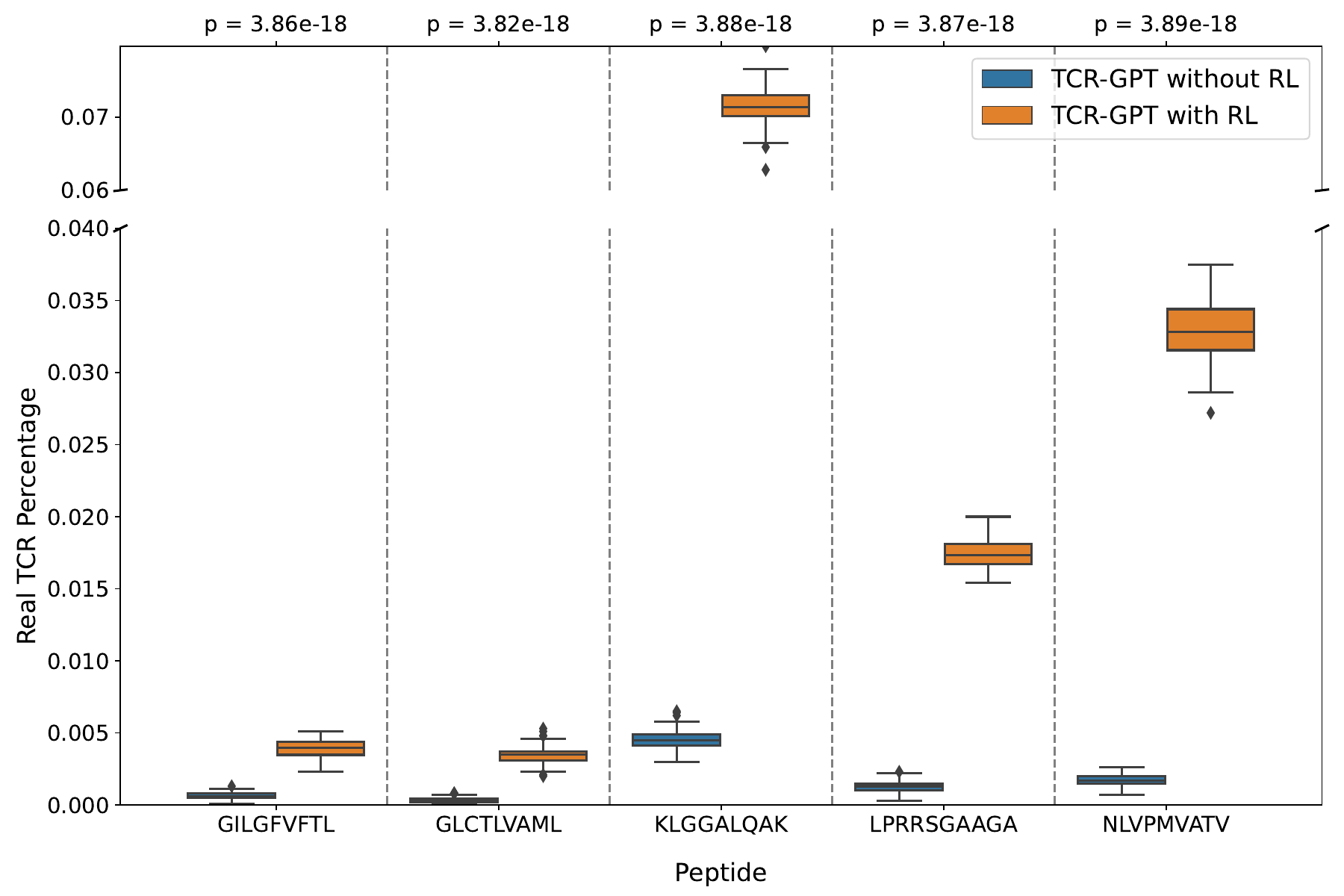}
    \captionof{figure}{Comparison of the percentage of TCRs generated from TCR-GPT overlapped with real TCRs binding to specific peptides, with and without RL fine-tuning.}
    \label{fig7}
\end{strip}

\subsection{Generating peptide-specific TCRs using TCR-GPT fine-tuned with RL}
Using TCR-GPT trained with universal TCR sequences, we address a more practical scenario. For a specific peptide, we employ RL to fine-tune the learned distribution, adapting it from universal TCR sequences to those that can bind to the specific peptide(Methods). PanPep
 \cite{ref_07_Gao}
 can output binding score ranges from 0 to 1 that indicate the binding probability of the given peptide and TCR. Based on this, we defined the binding percentage as the proportion of peptides generated by TCR-GPT with RL that have a binding score from PanPep exceeding threshold 0.5. RL demonstrated promising training efficiency, as evidenced by the significant increase in binding percentage of five tested peptides shown in Figure \ref{fig6}.

RL adjusts the distribution of learned TCR sequences, and it is crucial to ensure that the tuned model can generate TCR sequences similar to actual TCRs that bind to the target peptide. 
To verify this, we assess how many TCR sequences generated by the model—with and without RL—match real TCR sequences known to bind the target peptide. 
For each tested peptide, we compiled a set of known TCRs from IEDB \cite{ref_20_Vita2018} that can bind to it. 
We then used both models to generate 10,000 TCR sequences each and calculated the proportion of these sequences that matched the known TCRs. 
This process was repeated 100 times, with each proportion serving as a sample. 
we observed that the proportions of generated TCR sequences overlapped with real TCRs capable of binding to the given peptide were significantly higher for the TCR-GPT model fine-tuned with RL compared to the one without RL, as shown in Figure \ref{fig7}. 
This demonstrates that RL is capable of fine-tuning the TCR-GPT model to match the distribution of real TCRs specifically binding to the given peptide.

\section{Conclusion}
Text-generation models, driven by advanced natural language processing (NLP) techniques, have the potential to transform our understanding and utilization of TCR diversity. In our research, we introduce TCR-GPT, a sophisticated probabilistic model based on a decoder-only transformer architecture. This model is designed to identify sequence patterns within TCR repertoires and generate TCR sequences from the learned probability distribution. By implementing RL, we tailored the TCR sequence distribution to produce repertoires capable of recognizing specific peptides. This advancement holds significant potential for revolutionizing targeted immune therapies and vaccine development, paving the way for practical and effective medical applications.

The limitations of this study primarily stem from the fact that TCR-GPT is a probabilistic model focused exclusively on the CDR3 region of the TCR beta chain, thereby excluding the full-length sequences of both the alpha and beta chains of TCR sequences. This narrowed focus restricts the comprehensiveness of the model in representing the complete TCR repertoire. Future research efforts will be directed towards collecting paired full-length TCR sequences and scaling up the model’s parameters to accommodate the need for constructing a robust probability distribution of the full-length TCRs. Despite these current limitations, we are confident in the potential of the TCR-GPT architecture to effectively address these challenges with an adequately sized dataset. Additionally, we believe that RL will continue to be a highly efficient method for fine-tuning the distribution to meet specific targets, thereby enhancing the applicability and precision of our model in practical scenarios.

\bibliography{aaai25}

\end{document}